\documentclass[sigconf]{acmart}

\AtBeginDocument{%
  }

\setcopyright{acmlicensed}
\copyrightyear{2018}
\acmYear{2018}
\acmDOI{XXXXXXX.XXXXXXX}

\acmConference[Conference acronym 'XX]{Make sure to enter the correct
  conference title from your rights confirmation emai}{June 03--05,
  2018}{Woodstock, NY}
\acmISBN{978-1-4503-XXXX-X/18/06}

\begin{document}

\title{Transformer-based Single-Cell Language Model: A Survey}

\author{Wei Lan}
\email{lanwei@gxu.edu.cn}
\orcid{1234-5678-9012}
\author{Guohang He}
\email{18775067872@163.com}
\affiliation{%
  \institution{The Guangxi Key Laboratory of Multimedia Communications and Network Technology, School of Computer, Electronic and Information, Guangxi University}
  \city{Nanning}
  \country{China}
}

\author{Mingyang Liu}
\affiliation{%
  \institution{The School of Computer and Electronic Information, Guangxi University}
  \city{Nanning}
  \country{China}}
\email{hitomil@foxmail.com}

\author{Qingfeng Chen}
\authornote{Both authors contributed equally to this research.}
\affiliation{%
  \institution{The School of Computer, Electronic and Information, Guangxi University}
  \city{Nanning}
  \country{China}
}
\email{qingfeng@gxu.edu.cn}

\author{Junyue Cao}
\authornotemark[1]
\affiliation{%
  \institution{The College of Life Science and Technology, Guangxi University}
  \city{Nanning}
  \country{China}
}
\email{qingfeng@gxu.edu.cn}

\author{Wei Peng}
\affiliation{%
  \institution{the Faculty of Information Engineering and Automation, Kunming University of Science and Technology}
  \city{Kunming}
  \country{China}}
\email{weipeng1980@gmail.com}

\renewcommand{\shortauthors}{Lan et al.}

\begin{abstract}
The transformers have achieved significant accomplishments in the natural language processing as its outstanding parallel processing capabilities and highly flexible attention mechanism. In addition, increasing studies based on transformers have been proposed to model single-cell data. In this review, we attempt to systematically summarize the single-cell language models and applications based on transformers. First, we provide a detailed introduction about the structure and principles of transformers. Then, we review the single-cell language models and large language models for single-cell data analysis. Moreover, we explore the datasets and applications of single-cell language models in downstream tasks such as batch correction, cell clustering, cell type annotation, gene regulatory network inference and perturbation response. Further, we discuss the challenges of single-cell language models and provide promising research directions. We hope this review will serve as an up-to-date reference for researchers interested in the direction of single-cell language models.
\end{abstract}



\maketitle

\section{Introduction}
\label{s:introduction}
\noindent
Single-cell research has shown tremendous potential across a variety of fields including genetics, immunology and oncology. By utilizing single-cell RNA sequencing data for cluster analysis and the identification of cell subtypes, it is possible to accurately categorize cell populations and reveal crucial information about cell interactions and the structure of tissues \cite{lan2023multiview}. Exploring  the gene expression, gene function and gene-gene interaction at the single-cell level helps to unveil the deep mechanisms of cellular heterogeneity within tissues \cite{unveiling,landeep}. Single-cell research is critically important for understanding fundamental biological processes and provides significant insights for the diagnosis of diseases \cite{12}. Single-cell data usually consist of large amounts of high-dimensional data which contains complex information. There is heterogeneity among single-cell data originating from the same tissue.

In the early stages, traditional machine learning methods, such as n-gram \cite{ngram} and Hidden Markov Models (HMM) \cite{hmm}, were widely used for cell annotation and protein prediction. With the development of machine learning technology, more sophisticated algorithms were applied to single-cell research \cite{machine}. Subsequently, deep learning models, including Recurrent Neural Networks (RNN) \cite{3} and Convolutional Neural Networks (CNN) \cite{4}, were used for the analysis of single-cell data. Currently, the transformers developed by Google has become the most popular language model \cite{10}. The transformers can process an entire sentence at once during training and effectively captures long-distance dependencies within sequences through the self-attention mechanism \cite{41}. This capability enables transformers to effectively explore various types of single-cell data. It leads to an increasing number of researchers applying Transformer technology in the field of single-cell research \cite{osnet}. 

This review will introduce the main modules of the transformers in the second section. Then, we provide an overview and analysis of existing single-cell language models in the third section and showcase some downstream tasks accomplished by single-cell language models in the fourth section. Final, we  discuss the challenges and opportunities of transformers-based single-cell language models in the fifth section. We hope to offer assistance to individuals interested in understanding single-cell language models.

\begin{figure*}[t]
\centerline{\includegraphics[scale=0.6]{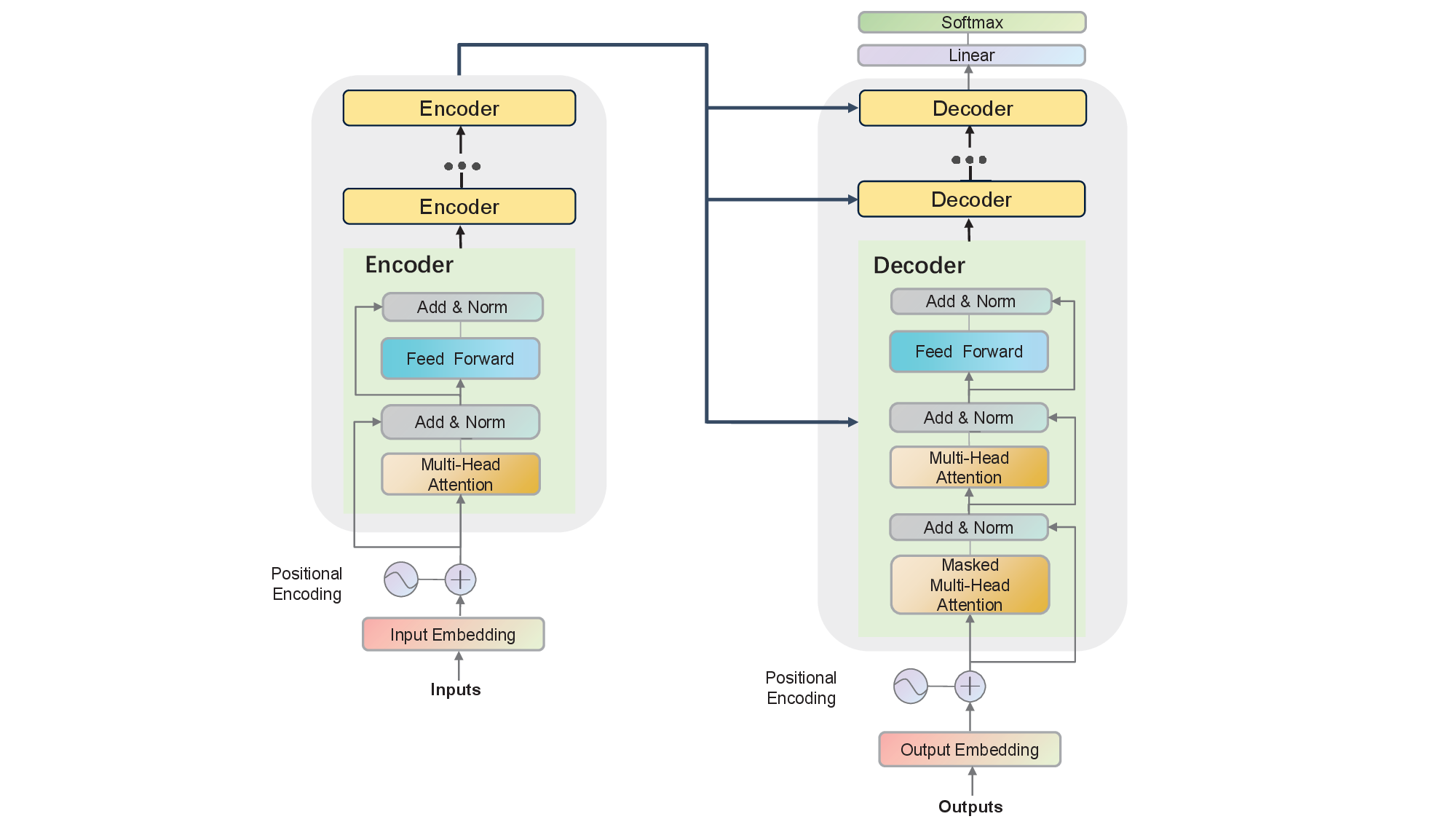}}
\caption{The structure of transformers}
\label{Transformer}
\end{figure*}

\section{Transformer}
\label{s:Experimental}
\noindent
The transformers requires extensive training on numerous texts. It usually employs a self-supervised approach during training, enabling language models to perform classification and generation \cite{25}. For instance,the transformers-based language models can automatically extract key information of text, generate new text and answer user queries in question-answering. These achievement is credited to the ability of transformers for learning long-term dependencies of language and allowing parallel training across multiple language units. This enhances the parallelism in processing sentences and capability to extract overall sequence correlations of transformers. The structure of transformers is depicted in Fig. 1.

The transformers has demonstrated excellent performance in both training tasks from scratch and pre-training tasks. Transformer-XL \cite{20} introduces the recursive mechanism and positional encoding. It captures longer-term dependencies by learning beyond fixed-length dependencies while maintaining temporal continuity to address context fragmentation. Reformer \cite{23} reduces attention calculation complexity and uses reversible residual layers instead of standard residual layers to achieve higher memory efficiency and alleviate pressure on computing resources. In addition, pre-training tasks can reduce dependence on annotated data, thus lowering the training cost of the transformers \cite{14}.  The GPT \cite{15} employs multiple layers of the transformers encoders and performs unsupervised language modeling tasks during pre-training to learn semantic and syntactic knowledge from the text. The BERT \cite{16} is a model that pre-trained on large datasets. It use bi-directional transformers and mask mechanism to consider the context information from both the left and right sides of the input sequence simultaneously. Due to the success of these models, many models based on them have started to emerge. XLNet \cite{19} is a pre-training model base on Transformer-XL that achieves bidirectional learning of context. It uses the self-regressive strategy to helps the model avoid the inconsistency issue in pre-training fine-tuning. RoBERTa \cite{17} is a model based on Bert and achieves enhanced training performance by utilizing dynamic masking. 

\subsection{Encoder and decoder}
\noindent
The transformers is primarily composed of encoders and decoders, which uses residual connections and layer normalization. The layer normalization is defined as follows:
\begin{equation}
\mathrm{LayerNorm}(X+\mathrm{MultiHead}(X))
\end{equation}
\begin{equation}
\mathrm{LayerNorm}(X+\mathrm{FFN}(X))
\end{equation}
where the X in formula(1) denotes the input embedding. It is processed through multi-head self-attention mechanism(MultiHead). After processing X, the result is added to the original X in formula(1) to obtain the X in formula(2). Then the X in formula(2) is processed through Position-wise feed-forward networks(FFN). The Layer normalization computes the mean and variance of each input sequence to provide more accurate training results \cite{30}. The encoder gradually extracts semantic information from the input sequence and encodes it into a series of hidden vectors by stacking multiple identical layers. The decoder is responsible for transforming the hidden representations generated by the encoder into an output sequence. It adopts an autoregressive training approach. The decoder acquires information about the entire sequence of tokens during training, which would lead to a decrease in prediction accuracy. To address this issue, the decoder uses masked self-attention mechanism in the first layer. After obtaining vector information based on the masked self-attention mechanism, it needs to be combined with the hidden vectors provided by the encoder before entering the next layer. Then, the decoder gradually generates vectors of the sequence and transforms them into the final output sequence based on linear transformation and Softmax function.

\subsection{Multi-head self-attention mechanism}
\noindent
The multi-head self-attention mechanism is comprised of multiple self-attention mechanism. It can help the model to determine the important of parameters during the training process. In addition, it adjusts the weights at different positions by calculating the correlations between each input position and other positions. 
The self-attention mechanism is defined as follows:

\begin{equation}
\mathrm{Attention}(Q,K,V)=\mathrm{softmax}(\frac{\mathrm{QK^T}}{\sqrt{d_\mathrm{k}}})V
\end{equation}
where $d_\mathrm{k}$ represents the dimensionality of the key vector. $Q$, $K$ and $V$ are three matrices. $K^\mathrm{T}$ represents the transpose of the $K$ matrix. The dot product of $Q$ and $K^\mathrm{T}$ denotes the similarity between the current word vector and other word vectors. After dividing this value by $\sqrt d_\mathrm{k}$ and applying the softmax function, the coefficient of weight is obtained. The weight coefficient is then multiplied by V to ultimately obtain the attention value. The multi-head attention mechanism is defined as follows:
\begin{equation}
\mathrm{M}(Q, K, V)=\mathrm{C}(\mathrm{head_1},\ldots,\mathrm{head_h})W^\mathrm{O}
\end{equation}
\begin{equation}
\mathrm{head_i}=\mathrm{Attention}(\mathrm{QW_i^{Q}},\mathrm{KW_i^K},\mathrm{VW_i^V})
\end{equation}
where $W^\mathrm{O}$ is a matrix containing the weights for each attention value. C denotes the concat function. $\mathrm{head_i}$ represents the self-attention mechanism module of $\mathrm{i}$-th head. $W^\mathrm{O}$ contains the weights of each $\mathrm{head_i}$. The $W^\mathrm{Q}$, $W^\mathrm{K}$ and $W^\mathrm{V}$ denote the weight matrices. Each input embedding vector is multiplied with them to obtain the corresponding $Q$, $K$ and $V$ matrices. They are updated with each backward propagation during training. Each self-attention module has different $W^\mathrm{Q}$, $W^\mathrm{K}$ and $W^\mathrm{V}$. The multi-head attention value is calculated by weighting each Attention value with $W^\mathrm{O}$.

\subsection{Position encoding}
\noindent
The position encoding is obtained by adding positional information to the embedding vectors of input words in transformers. It is defined as follows:
\begin{equation}
\mathrm{PE}_{(\mathrm{pos},\mathrm{2i})}=\mathrm{sin(pos/{10000}}^{2i/d_\mathrm{{model}})}
\end{equation}
\begin{equation}
\mathrm{PE}_{(\mathrm{pos},\mathrm{2i+1})}=\mathrm{cos(pos/{10000}}^{2i/d_\mathrm{{model}})}
\end{equation}
where pos is the position index, i is the dimension index and $d_{\mathrm{model}}$ is the size of the hidden layer. The sine and cosine values for each pos and i are calculated separately using the PE function. Then they are merged into a position encoding vector. This ensures that the embedding vectors for each token not only contain semantic information, but also position information of input sequence. In addition, the relative position encoding is proposed \cite{35}. It makes transformers to better understand the positional information of input sequence, thereby enhancing the performance and generalization capability of model.

\subsection{Position-wise feed-forward networks}
\noindent
The position-wise feed-forward networks(FFN) acts as a multi-layer perceptron, which is equivalent to use a linear layer in each encoder and decoder \cite{104}. It is defined as follows:
\begin{equation}
\mathrm{FFN}(x)=\max{(0,xW_1+b_1)}W_2+b_2
\end{equation}
where $W_1$, $b_1$, $W_2$ and $b_2$ are parameters that can be learned during training. The FFN initially performs a linear operation on the input to increase its dimension and applies the ReLU activation function to learn more complex feature information. In final, the FFN reduces the dimension to the original dimension based on a linear operation. This contributes to enhance the generalization capability of features.

\section{The Application of Transformer in Single-cell}
\noindent
We categorize the application of single-cell data analysis based on transformers into single-cell language models and single-cell large language models depended on whether it uses pre-training or not. These models effectively analyze single-cell omics data by utilizing the unique feature representation of transformers.

\subsection{single-cell language model}
\noindent
This section introduces the current structural design and optimization of single-cell language models. These models are developed based on the transformers framework. They have been utilized for analyzing various types of single-cell datasets, including single-cell transcriptomics, spatial transcriptomics, epigenomics.

\subsubsection{Single-cell language model based on single-cell transcriptomics}
\noindent
The transCluster \cite{65} is a model based on transformers for analyzing scRNA-Seq data. It demonstrates that transformers can be used for scRNA-seq analysis. It utilizes Linear Discriminant Analysis (LDA) \cite{125} to obtain input embeddings for the transformers. Then, CNN is employed to train the output of transformers for predicting cell types. In addition, scTransSort \cite{scTransSort} is also a model that combination of transformers and CNN. It uses CNN to transfer the gene embeddings of each cell into multiple two-dimensional matrix blocks. Each matrix block represents a token and these tokens are trained through 12 layers of transformers. Finally, a linear classifier utilizes the output features of transformers to predict cell type. CIForm \cite{66} is a model inspired by the application of transformers in computer vision (CV). It divides equally sized sub-vectors within the gene embedding module. These sub-vectors are combined with positional embeddings to fed into the transformers for training. The average pooling layer uses the average of the output sub-vectors to Fetch the final result. STGRNS \cite{101} is an interpretable model base on transformers. It proposes a Gene Expression Motif (GEM) data processing technique to process scRNA-seq. The combination of GEM and transformers in STGRNS provides stonger interpretability. In contrast to STGRNS, T-GEM \cite{116} enhances model interpretability by replacing the weights in the transformers with gene-related weights. It obtains attention values for different genes. Then, it utilizes these attention values for the classification task.

\subsubsection{Single-cell language model based on single-cell spatial omics and epigenomics}
\noindent
The PROTRAIT \cite{67} is a model based on transformers for analyzing scATAC-Seq data. It utilizes one-hot encoding to map input sequences into a latent space. When the sequence length is less than a predefined threshold, the one-hot encoding is transformed into motif embedding through convolutional layers. For sequences longer than the predefined threshold, an alternating combination of convolutional and pooling layers is used to obtain motif embedding. Then, the embeddings with absolute positional information are subsequently passed into the transformers for further processing. The output features from the transformers are used to conduct cell classification. TransformerST \cite{transformerST} constructs a Variational-Transformers framework for data representation and employs CNN as both the decoder and encoder. It introduces a graph transformers between the decoder and encoder to analyze spatial transcriptomics data. By constructing an undirected graph, the graph transformers is able to learn nonlinear mappings and aggregate neighbor relationships. It makes high-resolution reconstruction of gene expression possible.

\subsubsection{Single-cell language model based on single-cell multi-omics}
\noindent
The SCMVP \cite{scMVP} is a deep generative model based on transformers specifically designed for the simultaneous analysis of scRNA-seq and scATAC-seq data. The model establishes two independent channels at the encoder and decoder layers for processing scRNA data and scATAC data. In the scRNA channel, the masked attention mechanism is adopted, while in the scATAC channel, the self-attention mechanism is employed. Subsequently, the outputs of the two channels are combined, and the mean and variance of the common latent variables are obtained through a shared linear layer.
scMoFormer \cite{scMoFormer} is a multimodal model based on transformers that uses a heterogeneous graph to model single-cell data. It constructs a multimodal heterogeneous graph containing three types of nodes: cells, genes, and proteins. In the training framework, three transformers are used, each dedicated to extracting the data representation of the corresponding modality. Finally, a multi-layer fully connected network is utilized to predict the target protein expression level of each cell.
DeepMAPS \cite{113} is a model that introduces the heterogeneous graph transformer (HGT) framework. It constructs a heterogeneous graph using a cell-gene matrix. Then, the entire heterogeneous graph is divided into multiple subgraphs and HGT is applied on these subgraphs. Subgraph sampling is performed through a sparse-based feature selection method. During training process, the information of nodes is updated through multiple iterations of training and the training on different subgraphs shares the same set of parameters. After training on all subgraphs is completed, HGT is applied to the entire heterogeneous graph to obtain data features.
MarsGT \cite{marsgt} is an extended model based on DeepMAPS. The heterogeneous graph of MarsGT is constructed base on cell-gene matrix, gene-peak matrix and cell-peak matrix. Compared to DeepMAPS, it better obtain features of single-cell data from the perspective of regulatory networks by increasing the peak. During the subgraph sampling stage, a probability-based subgraph sampling method is employed to select genes and regulatory regions associated with rare cells. Then, the model is trained on the subgraph using transformers. After obtaining the trained weights, the pre-trained model is applied to the entire graph for training.

\begin{figure*}[p]
\centerline{\includegraphics[scale=0.6]{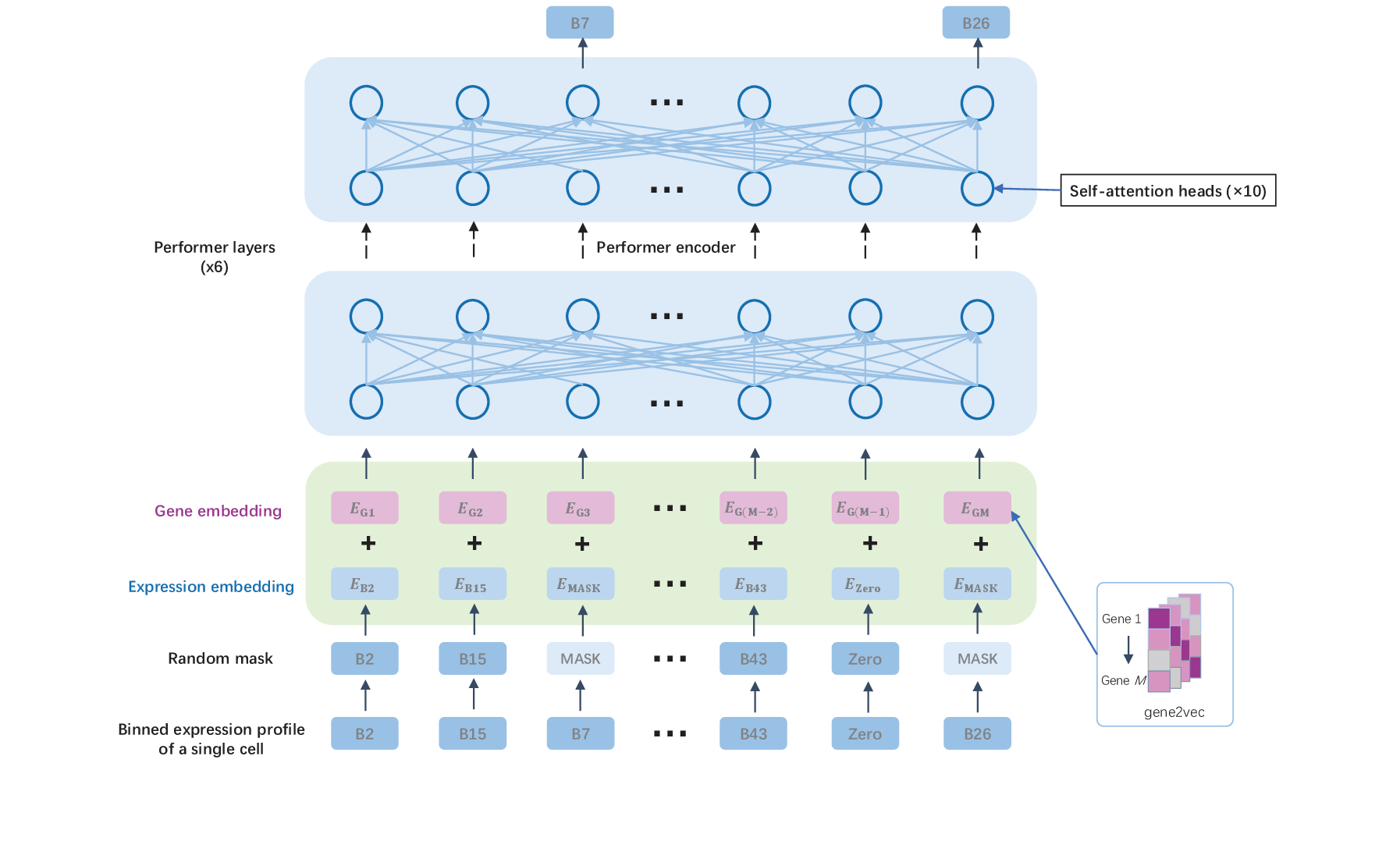}}
\caption{The framework of scBERT: B represents a discretized expression converted from scRNA seq data and is randomly masked; $E_\mathrm{B}$ represents the expression embedding of each gene; $E_\mathrm{G}$ is obtained by gene2vec and represents the gene embedding of each gene.}
\label{scBERT}
\vspace{20pt}
\centerline{\includegraphics[scale=0.6]{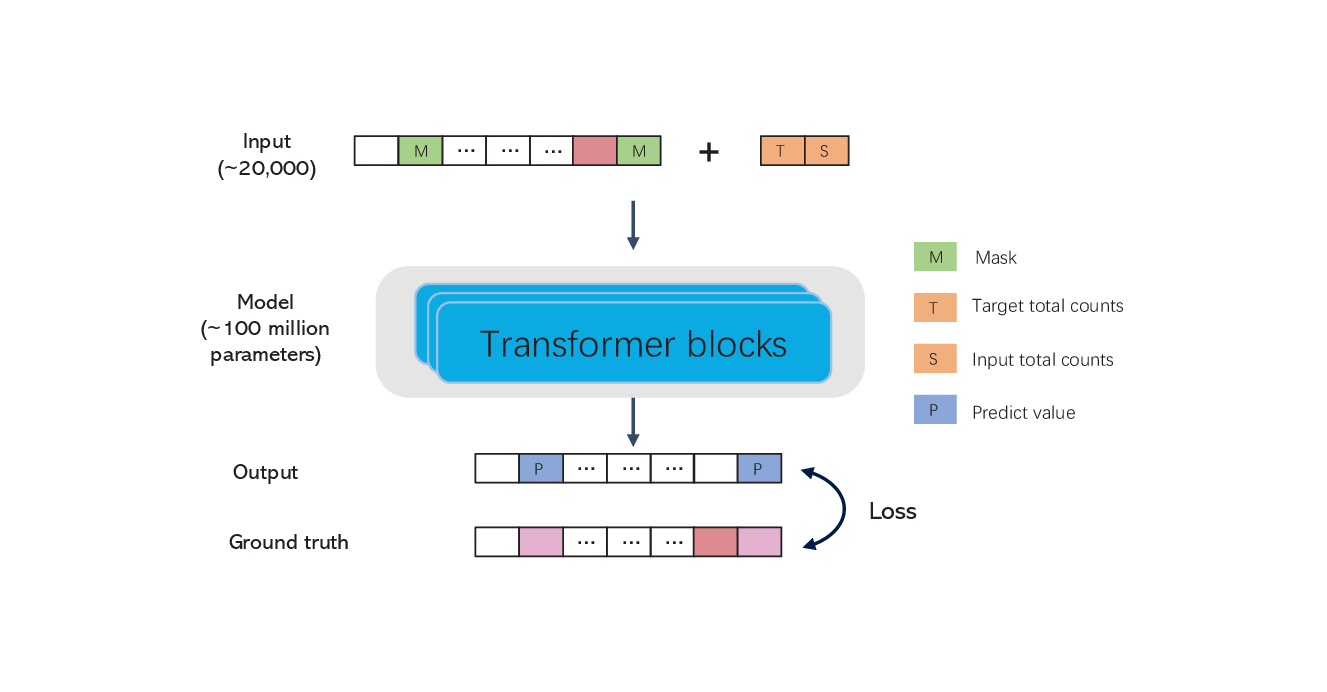}}
\caption{Pre-training Module Structure of scFoundation: the input consists of the masked gene expression vector and two total count indicators (T and S). The output is the predicted expression value for all genes.}
\label{scFound}
\end{figure*}

\subsection{Single-cell Large Language model}
\noindent
Currently, large language models are also being applied to single-cell domains. The GPT and BERT have emerged as leading representatives. This section will provide an introduction of the current single-cell large language models.

\subsubsection{Single-cell large language model based on single-Cell transcriptomics}
\noindent
The scBERT \cite{69} is the first single-cell pre-training model constructed based on the BERT architecture. The structure of scBERT is shown in Fig. 2. During the training process, scBERT has been optimized to eliminate of artificial biases and overfitting for enhancing the generalization capability of model. To capture the similarity between genes, the scBERT employs the gene2vec \cite{71} to obtain gene embedding for each gene. The input embedding information is obtained to capture relationships between genes by combining expression embedding and gene embedding. The embedding design allows scBERT to more effectively transform gene expression information into the input for the transformers for generating cell-specific embedding. Considering that most scRNA-seq data dimensions exceed the 512-limitation of transformers, scBERT utilizes the Performer to reduce computational complexity through approximate self-attention calculations, which employs a Linear Attention mechanism based on low-rank random feature mapping. It enables scBERT to input over 16,000 genes when processing long sequence data. In addition, the scBERT also provides the interpretability by using  Enrichr to visual attention weight to reflect the contribution of genes.

The scFoundation \cite{90} is a large pre-trained model based on transformers with 100 million parameter scale. The embedding module of scFoundation is employed to get final embeddings with positional information. In addition, scFoundation adopts an asymmetric encoder-decoder architecture. During the encode phase, it exclusively conducts the training on non-zero and non-masked expressed genes to reduce computational costs. In the decode phase,  it restores zero and masked expressed genes to learn relationships among all genes. The Read-depth-aware task is utilized as training strategy to train a pre-trained model, which is illustrated in Fig. 3. It successfully harmonize read depth differences across different cells to prove more coordinated and precise when dealing with cells with varying sequencing depths.  

\subsubsection{Single-cell large language model based on single-cell multi-omics}
\noindent
The scGPT \cite{89} is the first single-cell foundation model based on transformers that undergone generative pre-training on over 33 million cells. The model draws inspiration from GPT. The structure of scGPT is depicted in Fig. 4. scGPT treats genes as tokens and uses a condition token to represent the positional information of genes. In addition, it employs value binning to address differences between different sequencing batches. scGPT uses stacked transformers layers and Flash-Attention \cite{97} to handle single-cell multi-omics data. Flash-Attention can effectively address the sequence length limitation and reduce computational cost. In terms of interpretability, scGPT focusing on key genes through pre-training on a good deal of single-cell data. Therefore, it has more comprehensive interpretability. While scGPT demonstrates impressive performance, It still has some shortcomings. It proves competitive in low-data settings, but it requires careful consideration of experimental conditions in zero-shot settings. Moreover, the current pre-training methods may lack universal applicability. 

The CellPLM \cite{80} is the first single-cell pre-trained model based on transformers that considers the relationship between cells. The structure of CellPLM is depicted in Fig. 5. It establishes a gene expression embedder for processing input data. The embedder initializes an embedding vector for each type of gene and filters out unmeasured genes and randomly masked genes. The gene expression embedder aggregates gene embedding based on their expression levels in each cell and then transforms them into a suitable input of the transformers. These expression embedding are then input into a structure of an encoder-decoder by utilizing a latent space between the encoder and decoder. The encoder part comprises N transformers blocks. However, the computational complexity of transformers exhibits quadratic growth which results in significant computational costs \cite{lan2023benchmarking}. CellPLM replaces the transformers with a variant called Flowformer \cite{82} to resolve the input constraints and computational complexity problems associated with the transformers. To more effectively capture cell-cell relationships and spatial positional information of individual cells, CellPLM incorporates spatial resolution transcriptome (SRT) data into the encoder for training. SRT data contains position embedding information. The position embedding are combined with expression embedding to obtain the final input embedding. In the latent space, a Gaussian mixture model is employed. The decoder employs several feedforward layers (FFLayers) to train latent space vectors and acquires the batch label of each cell by learning from the learnable lookup table.

\begin{figure*}[p]
\centerline{\includegraphics[scale=0.7]{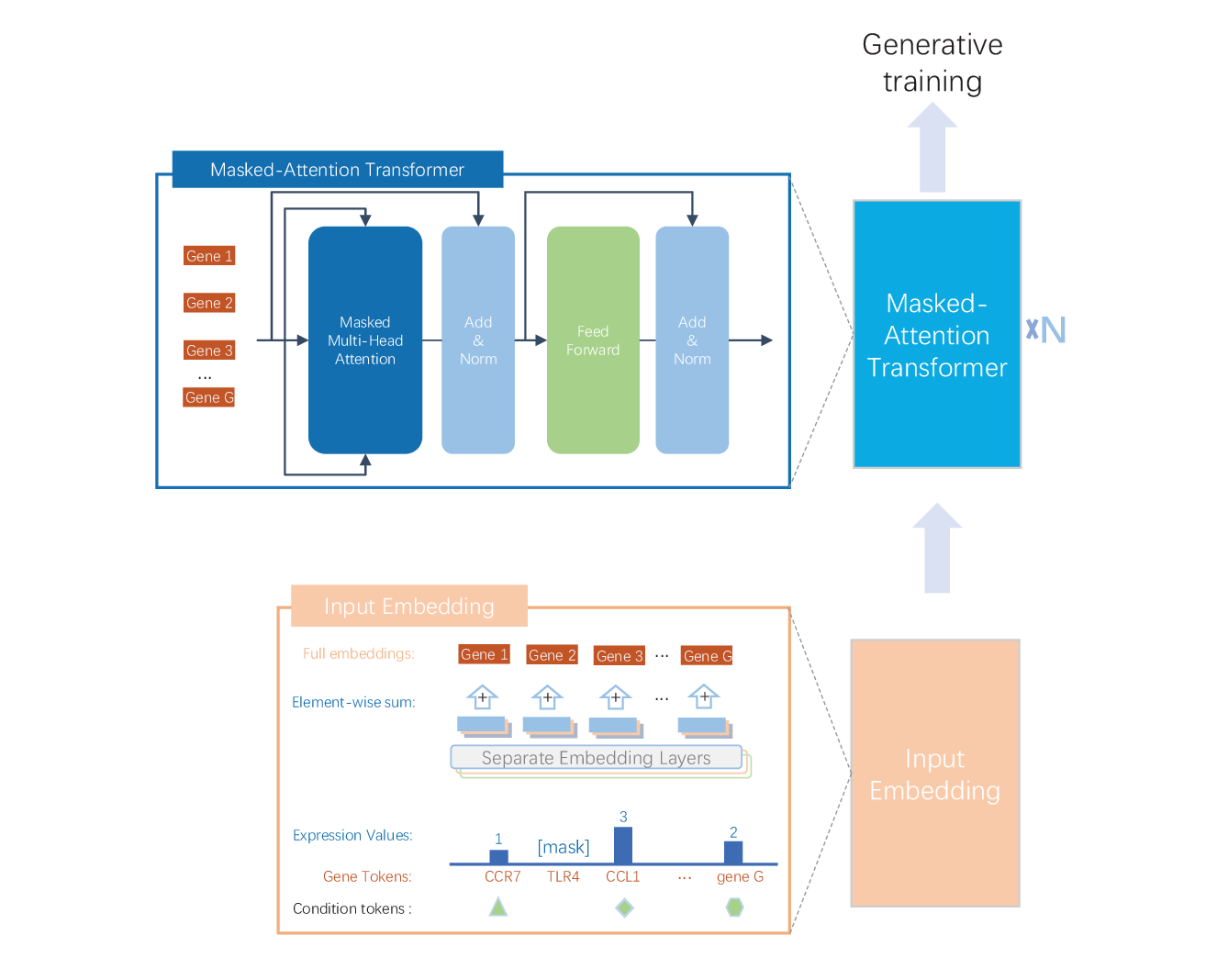}}
\caption{The framework of scGPT}
\label{scGPT}
\vspace{20pt}
\centerline{\includegraphics[scale=0.5]{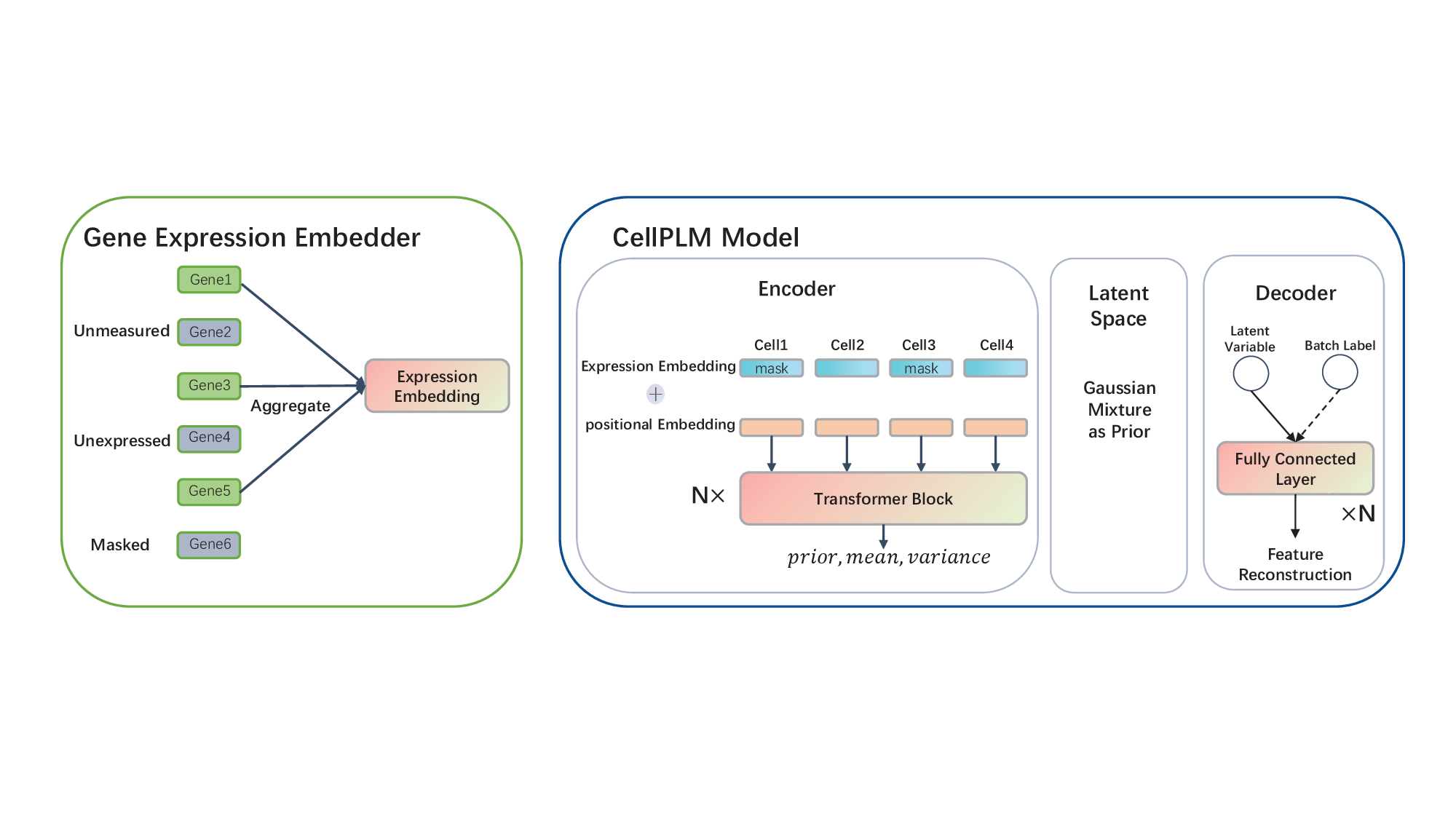}}
\caption{The training framework of CellPLM}
\label{CellPLM}
\end{figure*}

\subsubsection{Single-cell large language model based on gene expression ranking}
\noindent
The tGPT \cite{50} is an autoregressive unsupervised training model based on transformers. It utilizes the ranking of gene expression to predict the index of the next gene. Gene expression ranking provides the relative position of genes and is more suitable for large-scale gene screening and comparative analysis. However, this strategy may only consider genes with higher expression levels and neglect the specific information contained in low-expression genes. The structure of tGPT is depicted in Fig. 6. The tGPT predefines a length limit of input sequence and any part of the input sequence exceeding this limit is truncated, while the sections not reaching the limit are padded as 0. In the training process, it combines gene token embedding with positional encoding embedding. Final embedding undergoes 8 transformers modules to extract features from single-cell sequences.

\begin{figure*}[t]
\centerline{\includegraphics[scale=0.7]{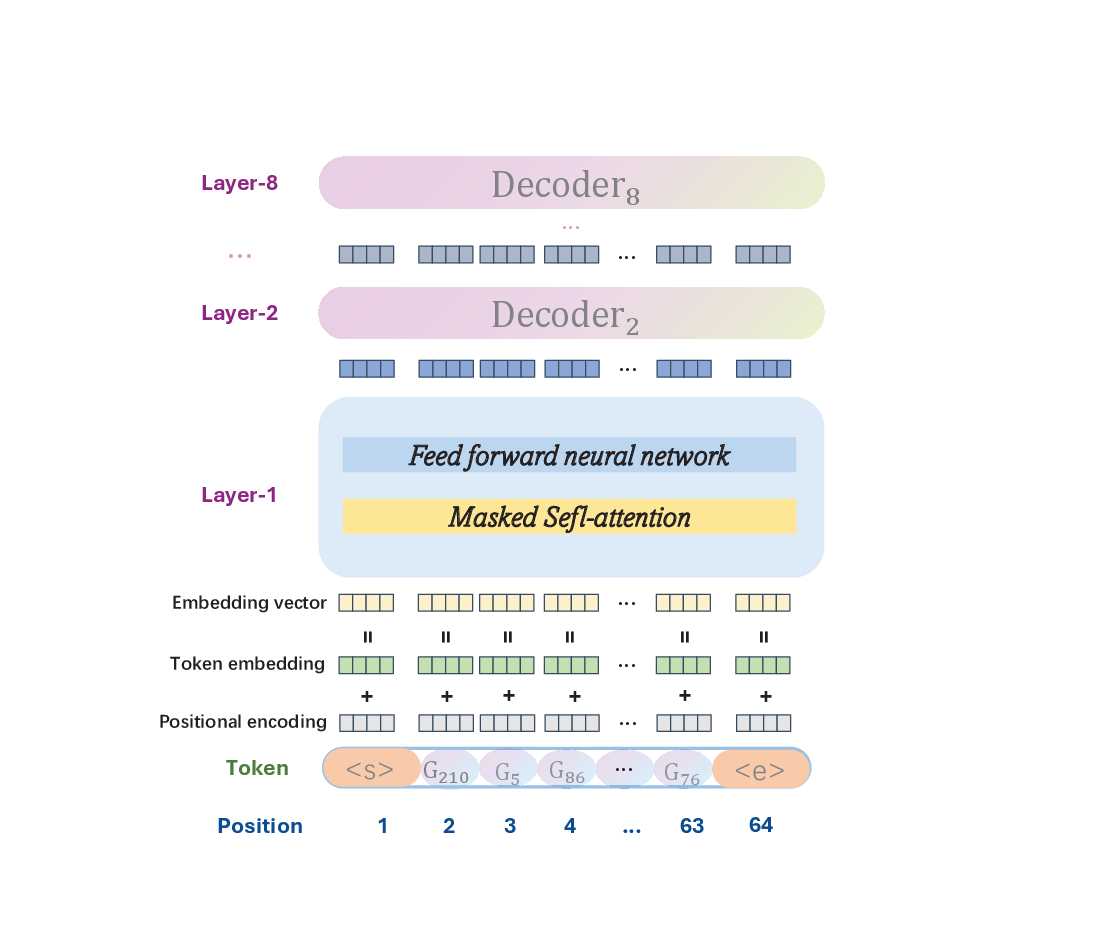}}
\caption{The module structure of tGPT}
\label{tGPT}
\end{figure*}

The Cell2Sentence (C2S) \cite{121} is a pretrained model fine-tuned on GPT-2, focusing on handling text sequences containing gene names. Through fine-tuning, C2S is capable of generating new cell sentences and reversely converting them back into gene expression vectors, retaining most of the information. The order of gene names is determined by the expression ranking of each gene and C2S uses these gene name sequences as its input. By converting cell text sequences back into gene expressions, C2S minimizes information loss and retains key information from the original data in most cases. This method enables transformers to acquire information about single-cell data, but the sequence conversion operation often results in higher computational costs.

\section{Downstream task analysis}
\noindent
The single-cell language models based on transformers have conducted on various downstream tasks include batch correction, cell clustering, cell type annotation, Gene network inference and Perturbation responses. The datasets used for these downstream tasks are primarily obtained through databases such as TCGA \cite{TCGA} and GEO \cite{geo}. The details of them are shown in Table 1.

\begin{table*}[t]
\centering
\caption{The details of downstream multi-task(Single-cell large language models are marked with an asterisk.)
}
\label{dataset}
\scalebox{0.8}{
\begin{tabular}{p{4cm}p{2cm}@{\hspace{5em}}p{5cm}@{\hspace{2em}}p{6cm}}
\hline
\textbf{Downstream tasks}& \textbf{Model}& \textbf{Metrics} & \textbf{Dataset}\\
\hline

Batch Correction & tGPT* & kBET & HCA\cite{137}\\
 & scGPT* & ASWbatch, GraphConn & COVID-19\cite{130}, PBMC 10\cite{131}, Perirhinal Cortex\cite{132}\\

Cell Clustering & scMVP & ARI & Paired-seq cell line data\cite{pair}, SNARE-seq cell line data\cite{SNARE-seq}\\
 & tGPT* & ARI, NMI, FMI & HCA\cite{137}, HCL\cite{134}, TCGA\cite{138}, Macaque Retina\cite{202}, the GTEx\cite{135}, Tabula Muris\cite{136} \\
 & CellPLM* & ARI, NMI & public dataset \cite{publicdata} \\
 & DeepMAPS & ASW, ARI & PBMC\cite{131}, lung tumor leukocytes CITE-seq dataset\cite{lung} \\

Cell Type Annotation & TransCluster & F1-score, Precision, Recall, MCC & Shao\cite{shao}, Baron\cite{baron}\\
 & PROTRAIT & ARI, AMI & sci-ATAC human atlas\cite{acat}\\
 & scBERT* & Accuracy, ARI, F1-score & Baron\cite{baron}, Muraro\cite{muraro}, Segerstolpe\cite{segerstolpe}, Xin\cite{xin},\\
 & scGPT* & Accuracy, Precision, Recall, F1-score & hPancreas\cite{139}, multiple sclerosis\cite{140}, tumor-infiltrating myeloid\cite{mye}\\
 & CellPLM* & Precision, F1-score & hPancreas\cite{139}, multiple sclerosis\cite{140}\\

Gene network inference & DeepMAPS & Closeness centrality, Eigenvector centrality, Functional enrichment analysis & Reactome\cite{reactome}, DoRothEA\cite{roroth}, TRRUST v2\cite{trrust}\\
 & scGPT* & Pathway enrichment analysis & Immune Human\cite{144}, ChIP-994Atlas database\cite{chip}, Adamson\cite{146} \\

Perturbation Prediction & scFoundation* & MSE &  Dixit\cite{147}, Adamson\cite{146}, Norman\cite{148}\\
 & scGPT* & PCC &  Adamson\cite{146}, Norman\cite{148}\\
 & CellPLM* & RMSE & Adamson\cite{146}, Norman\cite{148}\\

\hline
\end{tabular}
}
\end{table*}

\subsection{Batch correction}
\noindent
With the increasing quantity of single-cell data, the variability between different batches has become an increasingly significant interference in data analysis. It becomes an urgent challenge to improve the effectiveness of batch correction. Three key metrics are used to evaluation of batch correction effects including k-nearest neighbor batch effect test (kBET) \cite{kbet}, Average Silhouette Width for batch correction (ASWbatch) \cite{ASW} and Graph Connectivity measurement (GraphConn) \cite{graphconnectivity}. The kBET assesses the effectiveness of correction by comparing the distribution of cells within and between batches. Its acceptance rate reflects the uniformity of cell distribution after correction. A higher acceptance rate indicates the preservation of biological heterogeneity and a reduction in technical batch effects. The ASWbatch originates from the concept of silhouette width in cluster analysis. It is used to measure the clustering effect after removing batch effects. The GraphConn is a method for evaluating the connectivity between cells in the dataset after batch correction. It aims to quantify the enhancement of cell-to-cell connectivity post-correction for reflecting the reduction of batch effects.

The tGPT \cite{50} adopts the ranking of gene expression to void the interference of actual expressions of Highly Variable Genes (HVGs) and batch information during training. It is trained on the HCA dataset \cite{137}, utilizing the kBET acceptance rate to reflect the magnitude of differences between different batches. In addition, tGPT conducted an immune checkpoint blockade (ICB) clinical trial. By quantifying the expression features of different attention heads, it is demonstrated that these attention heads have prognostic significance in this clinical trial. scGPT \cite{89} conducts batch effect experiments by fine-tuning on pre-trained models. To quantify batch correction performance, scGPT calculates the average silhouette width ($ASW_{batch}$) and the graph connectivity measure (GraphConn) \cite{song2023benchmarking}. It computes the AvgBATCH (i.e average of ASWbatch and GraphConn,) to comprehensively represent batch performance. scGPT evaluates batch correction performance on three datasets including COVID-19 \cite{130}, PBMC 10 \cite{131} and Perirhinal Cortex \cite{132}. The evaluation is conducted against three methods including Seurat \cite{seruat}, Harmony \cite{ham} and scVI \cite{scvi}. scGPT achieves a best performance AvgBATCH value on the three datasets. However, scGPT does not achieve excellent batch effect correction in zero-shot settings \cite{110}.  

\subsection{Cell clustering}
\noindent
The goal of cell clustering analysis is to group cells based on their gene expression patterns. When evaluating the accuracy of clustering results, commonly used metrics include Adjusted Rand Index (ARI) \cite{ARI}, Average Silhouette Width (ASW) \cite{ASW} and Normalized Mutual Information (NMI) \cite{NMI}. ARI adjusts the Rand Index by comparing the observed pair-wise concordance to the expected random concordance and yields a measure of clustering consistency. ASW measures the difference in similarity between samples and different clusters by calculating the silhouette width for each sample. It offers a intuitive evaluation of clustering results. NMI utilizes normalized mutual information to eliminate the influence of the number of clusters and the total number of samples It makes it useful for comparing clustering results under different parameter settings.

The scMVP \cite{scMVP} employs a joint deep learning model to learn features from both scATAC data and scRNA data. It is trained on Paired-seq cell line data \cite{pair} and SNARE-seq cell line data \cite{SNARE-seq}. Then it utilizes UMAP visualization to perform cell clustering analysis on cell clusters. It successfully identifies different numbers of cell subpopulations and effectively separates the integration data of scRNA-seq and scATAC-seq. It confirms its effectiveness in cell clustering analysis. tGPT \cite{50} is applicable to large-scale tissue samples through pre-training. It partitions samples into distinct clusters that correspond to different organs. It is trained on six datasets including HCA \cite{137}, HCL \cite{134}, TCGA \cite{138}, Macaque Retina \cite{202}, GTEx \cite{135} and Tabula Muris \cite{136}. The experimental results demonstrate that it achieves excellent performance in cell clustering tasks. CellPLM \cite{80} conducts unsupervised clustering analysis by extracting cell embedding vectors from the dataset without fine-tuning. CellPLM achieves zero-shot clustering experiments on a public dataset \cite{publicdata}. It compares with PCA, Geneformer and scGPT. In the experiments, it achieves the highest ARI and NMI. DeepMAPS \cite{113} validates cell clustering on ten single-cell multi-omics datasets. It trains with 36 parameter combinations and compares with Seurat, MOFA+ \cite{mofa+}, TotalVI \cite{totalVI} and Harmony. In all experiments, DeepMAPS achieved the best ARI and ASW. Furthermore, DeepMAPS performs single-cell multi-omics integration analysis on the PBMC dataset \cite{131} and the CITE-seq dataset of lung tumor leukocytes \cite{lung}. It successfully identifies 13 cell types and validates its effectiveness.

\subsection{Cell type annotation}
\noindent
Cell type annotation refers to assigning known cell type labels to each cell or cell cluster, which aids in gaining a deeper understanding of the biological significance of the cells \cite{lan2020ldicdl}. When evaluating the performance of cell annotation, commonly used metrics include precision, recall, accuracy and the F1 score \cite{lan2024jlonmfsc}. Precision represents the proportion of true samples of a certain category among those predicted as that category by the model. Accuracy is the ratio of correctly classified samples to the total number of samples. Recall is the proportion of true samples of a certain category that the model correctly predicts as that category. The F1 score is the harmonic mean of precision and recall. It is used to comprehensively assess the performance of model.

The TransCluster \cite{65} is the first model to apply transformers to cell type annotation. It is trained on the Shao dataset \cite{shao} and the Baron dataset \cite{baron} and demonstrates efficient performance in cell type prediction tasks. PROTRAIT \cite{67} is trained on the sci-ATAC human atlas \cite{acat} and generates cell embeddings that reflect the distribution of scATAC-seq data.Then, it uses the k-nearest neighbors (KNN) for cell type annotation. scBERT \cite{69} is pre-trained on 9 scRNA-seq datasets. Then fine-tuning is performed on the trained model. Final, it uses the K-means algorithm to annotate cell types. scBERT performs cell annotation tasks on the Baron dataset \cite{baron}, the Muraro dataset \cite{muraro}, the Segerstolpe dataset \cite{segerstolpe} and the Xin dataset \cite{xin}. Both scGPT \cite{89} and CellPLM \cite{80} are trained on the hPancreas \cite{139} dataset and multiple sclerosis (MS) \cite{140} dataset to perform cell annotation task. scGPT performs normalization, log transformation and binning operations on gene expression values. Then cell type annotation is achieved through fine-tuning. In addition, scGPT is trained on the tumor-infiltrating myeloid dataset (Mye.) \cite{mye} and evaluated on query partitions of three previously unseen cancer types. The results indicate that scGPT has high accuracy in distinguishing immune cell subtypes. CellPLM adds a feedforward layer during the fine-tuning process and utilizes standard cross-entropy loss function for the fine-tuning process. Fine-tuned CellPLM exhibits a significant improvement in F1-score and precision metrics on two datasets compared to the from-scratch CellPLM.

\subsection{Gene network inference}
\noindent
Gene network inference analysis reveals regulatory associations between genes by comparing gene expression patterns under different conditions. Currently, single-cell language models based on transformers have introduced innovative perspectives to the study of gene regulatory networks \cite{lan2022detecting}. Centrality score metrics, including closeness centrality (CC) and eigenvector centrality (EC) \cite{centrality} is used to the experiment of single-cell language models. The CC assesses the average distance of a gene node relative to other gene nodes in the network. EC considers not only the number of connections of a gene node but also the importance of the other gene nodes that it is connected to. In addition, functional enrichment analysis \cite{functional} and pathway enrichment analysis \cite{path} are employed in experimental analysis. Functional enrichment analysis aims to identify biological functions or processes that are significantly enriched in a set of genes. Pathway enrichment analysis is similar to functional enrichment analysis but focuses more on known biochemical pathways \cite{lan2024lgcda}. It aims to deeply understand how genes function through synergistic interactions within specific biological pathways.

The DeepMAPS \cite{113} uses the Steiner Forest Problem (SFP) to identify genes contributing significantly to cell cluster features and constructs a gene correlation network. It defines sets of genes regulated by the same transcription factor (TF) as regulons and compares regulon activities between cell clusters. Then, it selects regulons with significantly higher activity scores as cell-type-specific regulons and constructs gene regulatory networks (GRNs) based on cell cluster regulons. After constructing GRNs, DeepMAPS conducted functional enrichment analysis. Specifically, it employs hypergeometric tests to compare the intersection of GRN results with regulons in the database and evaluates whether the predicted regulons in the GRN are enriched for the same functions or pathways as known regulons. DeepMAPS is trained on single-cell multi-omics datasets from the 10x database. The Experiment of DeepMAPS demonstrates that the GRNs exhibits a greater number of unique transcription factor (TF) and cell-type-specific regulons and they are enriched in specific functions or pathways. In addition, scGPT \cite{89} demonstrates high interpretability in gene regulatory network experiments. Pre-training enables scGPT to emphasize genes with intricate relationships. It improves the interpretability of scGPT. In the Human Leukocyte Antigen (HLA) dataset, scGPT forms a human leukocyte antigen (HLA) gene network through zero-shot learning. On the Immune Human dataset \cite{}, fine-tuned scGPT generates CD gene networks through zero-shot learning and visualization of the gene information. scGPT performs pathway enrichment analysis on the Reactome database \cite{reactome}. It successfully validates the extracted gene program and identifies 22 additional pathways. These experiments demonstrate the ability of scGPT to capture complex gene relationships. Through pre-training and fine-tuning, scGPT achieves stronger generalization capabilities.

\subsection{Perturbation responses}
\noindent
Single-cell perturbation prediction experiments aim to predict and analyze the biological responses of cells to external stimuli or changes introduced into single cells \cite{lan2022kgancda}. Mean Squared Error (MSE) and Root Mean Squared Error (RMSE) have become two important metrics for evaluating the performance of model in predicting how cells respond to specific perturbations \cite{landeep}. MSE is used to measure the accuracy of the model in predicting the response of single cells to specific perturbations. A lower MSE value indicates that the predictions of model are more consistent with the actual observed values. RMSE is the square root of MSE and provides an error measure in the same units as the original data. It directly reflects the magnitude of the prediction error.

In perturbation responses prediction experiments, scFoundation \cite{90} is combined with the GEARS \cite{129} to construct personalized gene co-expression graphs for each cell. It significantly improves the accuracy of gene perturbation predictions. It is evaluated on three datasets including the Dixit dataset \cite{147}, the Adamson dataset \cite{146} and the Norman dataset \cite{148}. It obtains lower Mean Squared Error (MSE) values. In addition, scGPT \cite{89} uses the pre-trained the parameters of embedding and transformer layers to initialize fine-tuning. The fine-tuning process uses genes with zero and non-zero expression. scGPT is compared with GEARS and CPA \cite{151} on the Adamson dataset and the Norman dataset. It accurately predicted the expression changes of the top 20 Differentially Expressed (DE) genes in the datasets. During the fine-tuning process, CellPLM \cite{80} initializes other components except the decoder with pre-trained weights. CellPLM is compared with GEARS and scGen \cite{152} on the Adamson dataset and the Norman dataset. It conducts two types of experiments (single-gene perturbation and double-gene perturbation) on the Norman dataset and only single-gene perturbation on the Adamson dataset. In each experiment, CellPLM exhibited lower Root Mean Squared Error (RMSE) than GEARS and scGen.

\section{Challenges and prospects}
\noindent
In the field of single-cell research, transformers contribute to a deeper understanding of these vast and complex datasets. It enhances the simulation and comprehension of cellular processes. In this section, we discuss the challenges encountered by transformers-based single-cell models. We focus primarily on limitations in the transformer-based single-cell language model including handling long sequence data, overfitting risks in pre-training, computational requirement and interpretability. In addition, we also analyze some potential future research directions.

\subsection{Sequence data processing}
The transformers-based single-cell language models have strong representational capabilities on single-cell sequence data. However, single-cell sequence data often contains excessively long sequences \cite{lan2022sciac}. It leads to an exponential increase in the computational complexity of these models. In addition, single-cell data with long sequences may contain more complex gene relationships. Nevertheless, the self-attention mechanism of the transformers tends to capture dependencies between adjacent positions in the sequence. It may causes the model to ignore some key gene information. scBERT \cite{69} adopts a variant of transformers called Performer to solve the problem. scBERT used the low-rank attention mechanism of Performer to avoid over-focusing on dependencies between adjacent positions. When dealing with sparse DNA sequences, the attention mechanism of Performer may exhibit better robustness. Although Performer achieves good results, there are certain challenges in terms of data precision and sensitivity to model parameters due to the low-rank attention mechanism. In addition, the effectiveness of Performer is not always superior to that of the traditional transformers for different datasets and tasks. However, it is undeniable that using some variants of the transformers has brought new insights to the research of single-cell language models.

\subsection{Overfitting risks in pre-training}
Although transformers-based single-cell language models are increasingly inclined to adopt pre-training techniques, the analysis of these pre-trained models in terms of overfitting issues is relatively limited. The characteristic of single-cell data lies in its diversity of types and different types of single-cell data may vary significantly. It may lead to an imbalanced distribution of pre-training samples and potentially causing overfitting on smaller datasets. To address this issue, data augmentation techniques can be introduced into the pre-training. Currently, Generative Adversarial Networks (GAN) have shown promising results in the field of single-cell data augmentation \cite{Gan}. By using GAN to generate synthetic data samples that are similar to the original data, the diversity of the dataset can be effectively increased. It can mitigate the overfitting problems caused by data imbalance. In addition, interpolating and extrapolating between original single-cell data samples can also be considered. By using methods such as linear interpolation, polynomial interpolation or deep learning models to generate new samples, the quantity and diversity of the data can be increased. It further enhances the generalization capability and robustness of models. We believe that incorporating these methods into the pre-training process of single-cell language models may help address the issue of overfitting in the models.

\subsection{Computing Requirement}
Currently, transformers-based single-cell language models for single-cell multi-omics research are still in their early stages. Future work may involve incorporating more omics data in the pre-training phase to study single-cell multimodal tasks. However, the incorporation of omics data has led to an even larger scale of data. It causes challenges related to computational costs. Recently, the combination of recurrent neural networks and transformers has reduced computational costs by speeding up the training of transformers \cite{31}. This method could be considered as a possibility for application in single-cell language models. In addition, the parallel computing capabilities of transformers still face challenges. In the self-attention mechanism, the attention weights for each position need to be calculated sequentially and cannot be directly parallelized. When processing batch data, the sequence lengths of different single-cell samples may vary, increasing the complexity of parallel computing. In the future, solving the parallel computing capabilities of single-cell language models may become increasingly critical.

\begin{table*}[t]
\centering
\caption{The link to the code of the models.
}
\label{code}
\begin{tabular}{p{3cm}p{4cm}p{8cm}}
\hline
\textbf{Model} & \textbf{Input data type}& \textbf{Data repositories address} \\
\hline
TransCluster & scRNA-seq & https://github.com/ Danica123/TransCluster.git \\

scTransSort & scRNA-seq &https://github.com/ jiaojiao-123/scTransSort \\

CIForm & scRNA-seq &https://github.com/zhanglab-wbgcas/ CIForm\\

STGRNS  & scRNA-seq &https://github.com/zhanglab-wbgcas/ STGRNS \\

T-GEM & scRNA-seq, Transcriptomics(the pan-cancer RNA-Seq)&https://github.com/TingheZhang/TGEM\\

PROTRAIT  & scATAC-seq &https://github.com/ZhangLab312/PROTRAIT \\

scMVP  & scRNA-seq, scATAC-seq &https://github. com/bm2-lab/scMVP \\

scMoFormer  & scRNA-seq, Proteomics &https://github.com/OmicsML/scMoFormer \\

DeepMAPS  & scRNA-seq, scATAC-seq, CITE-seq &https://github.com/OSU-BMBL/deepmaps \\

MarsGT  & scRNA-seq, scATAC-seq &https://github.com/mtduan/marsgt\\

scBERT   & scRNA-seq & https://github.com/TencentAILabHealthcare/scBERT \\

scFoundation  & scRNA-seq &https://github.com/biomapresearch/scFoundation \\

scGPT   & scRNA-seq&https://github.com/bowang-lab/scGPT \\

CellPLM  & scRNA-seq, Spatial Transcriptomics, Perturb-seq & https://github.com/OmicsML/CellPLM \\

tGPT   & scRNA-seq & https://github.com/deeplearningplus/tGPT \\

Cell2Sentenc  & scRNA-seq &https://github.com/vandijklab/cell2sentence-ft \\
\hline
\end{tabular}
\end{table*}

\subsection{Interpretability}
Transformers-based single-cell language models offer significant advantages in terms of interpretability. They are capable of assigning different gene weights during the processing of sequence data to identify key features in the representation process. In single-cell research, this capability is crucial for understanding complex biological processes such as gene expression, protein interactions and gene regulation \cite{lan2024deepkegg}. In addition, single-cell data is highly complex and diverse. Each cell potentially exhibit unique gene expression patterns \cite{120}. Through the self-attention mechanism, transformers have successfully provided interpretability for the predictions of the key features. This helps biologists understand how models assign weights to different genes or cells and gain insights into gene expression patterns. Although transformers-based single-cell language models have achieved good results, these models still employ a black-box training approach. It inevitably affects the application of models in clinical settings. Therefore, improving the interpretability of single-cell language models remains a challenging research problem.

\subsection{Validation Analysis}
The single-cell language models and single-cell large language models mentioned in this paper have demonstrated promising results in experiments. Currently, some of these models have been subjected to benchmark experiments \cite{106,118,119}, which have revealed that different models exhibit varying performance across different tasks. These models have been proven to have the capability to integrate representations from diverse single-cell omics data. In particular, pre-trained models like scGPT have shown remarkable performance in gene function prediction tasks and achieve good results even without fine-tuning. However, the application of single-cell language models and single-cell large language models is still in its early stages and their generalizability faces certain challenges. In addition, comparing with some of the latest methods such as Sccross \cite{sccross}, ctpredictor \cite{ctpredictor}, will also help to promote research progress. Therefore, we provide an accessible link to the experimental code of the single-cell language model, please refer to Table 2 for details. We hope these resources can provide some assistance to researchers interested in this field.

\section{Conclusion}
\noindent
The Transformer-based Single-Cell Language Model has shown promising results in single-cell data analysis. In this review, we provide a detailed overview of single-cell language models and single-cell large language models. We summarized the methods of these models as well as their applications in downstream tasks. While these models may not achieve optimal performance in certain evaluation metrics, they hold potential contributions and applications in single-cell research. They open new possibilities for research and applications in the field and present significant avenues for further development. We think that the potential areas for improvement may include refining data preprocessing methods, reducing computational costs, enhancing model interpretability and optimizing the transfer learning process. In-depth investigations into these directions will facilitate more effective utilization of various types of single-cell data. This review aims to provide an overview of single-cell language models and hope promoting progress in the field of single-cell research.

\section*{Acknowledgements }
This work was partially supported by the National Natural Science Foundation of China (No. 62072124), the Natural Science Foundation of Guangxi (No. 2023JJG170006), the Natural Science and Technology Innovation Development Foundation of Guangxi University (No. 2022BZRC009), the CAAI-Huawei MindSpore Open Fund (No. CAAIXSJLJJ-2022-022A), the Project of Guangxi Key Laboratory of Eye Health (No. GXYJK-202407), the Project of Guangxi Health Commission eye and related diseases artificial intelligence screen technology key laboratory (No. GXYAI-202402).

\bibliographystyle{unsrt}
\bibliography{reference}

\end{document}